\documentclass{llncs} % For LaTeX2e
\usepackage{url}
\usepackage{graphicx,subcaption}              % to include figures
\usepackage{epstopdf}
\usepackage{amsmath}               % great math stuff
\usepackage{amsfonts}              % for blackboard bold, etc
\usepackage{verbatim}
\usepackage{notoccite}
\usepackage{array}
\usepackage{algorithm2e}
\title{NeuroHex: A Deep Q-learning Hex Agent}

\author{
Kenny Young, Gautham Vasan, Ryan Hayward
}
\institute{Department of Computing Science, University of Alberta}

\begin{document}

\maketitle
% Abstract
% Abstract
\begin{abstract}
DeepMind's recent spectacular success in using deep convolutional neural nets and machine learning to build superhuman level agents --- e.g.\ for Atari games via deep Q-learning and for the game of Go via Reinforcement Learning ---
raises many questions, including to what extent these methods will succeed in other domains. In this paper we consider DQL for the game of Hex: after supervised initializing, we use selfplay to train NeuroHex, an 11-layer CNN that plays Hex on the 13$\times$13 board. Hex is the classic two-player alternate-turn stone placement game played on a rhombus of hexagonal cells in which the winner is whomever connects their two opposing sides. Despite the large action and state space, our system trains a Q-network capable of strong play with no search. After two weeks of Q-learning, NeuroHex achieves win-rates of 20.4\% as first player and 2.1\% as second player against a 1-second/move version of MoHex, the current ICGA Olympiad Hex champion. Our data suggests further improvement might be possible with more training time.
\end{abstract}

%---------------------------------------- Introduction-------------------------------------------%
%\section{Introduction, Background and Motivation}
\section{Movitation, Introduction, Background}
\subsection{Motivation}
DeepMind's recent spectacular success in using deep convolutional neural nets and machine learning to build superhuman level agents --- e.g.\ for Atari games via deep Q-learning and for the game of Go via Reinforcement Learning ---
raises many questions, including to what extent these methods will succeed in other domains. 
Motivated by this success, we explore whether DQL can work
to build a strong network for the game of Hex.
\subsection{The Game of Hex}
%Hex is traditionally played on an 11x11 or 13x13 rhombus consisting of hexagonal cells (here we use the 13x13 variant). 
%Two players take turns placing stones of their color on the board, each trying to connect two different opposing edges (see figure \ref{fig:hex}). Due to the geometry of the board it is provably impossible for both players to connect their edges, and if the board is totally filled then one player is sure to be connected, thus no draws are possible. Despite the extremely simply rules, the game involves deep tactics and strategy and is one of the games used in the International Computer Games Association's annual Computer Olympiad. 
%RBH version:
Hex is the classic two-player connection game played on an $n$$\times$$n$
rhombus of hexagonal cells. Each player is assigned two opposite sides of the board and a set of colored stones; in alternating turns, each player puts one of their stones on an empty cell; the winner is whomever joins their two sides with a contiguous chain of their stones.  Draws are not possible (at most one player can have a winning chain, and if the game ends with the board full, then exactly one player will have such a chain), and
for each $n$$\times$$n$ board there exists a winning strategy for the 1st=player \cite{Gard57a}. Solving --- finding the win/loss value --- arbitrary Hex positions is P-Space complete \cite{Reis81}. 

Despite its simple rules, Hex has deep tactics and strategy. Hex has served as a test bed for algorithms in artificial intelligence since Shannon and E.F.\ Moore built a restistance network to play the game \cite{Shan53}. Computers have solved all 9$\times$9 1-move openings and two 10$\times$10 1-move openings, and 11$\times$11 and 13$\times$13 Hex are games of
the International Computer Games Association's annual Computer Olympiad \cite{HRpt13}.

In this paper we consider Hex on the 13$\times$13 board.
\begin{figure}[!ht]
\centering
\begin{subfigure}[t]{.45\textwidth}
  \centering
      \includegraphics[width=1\textwidth]{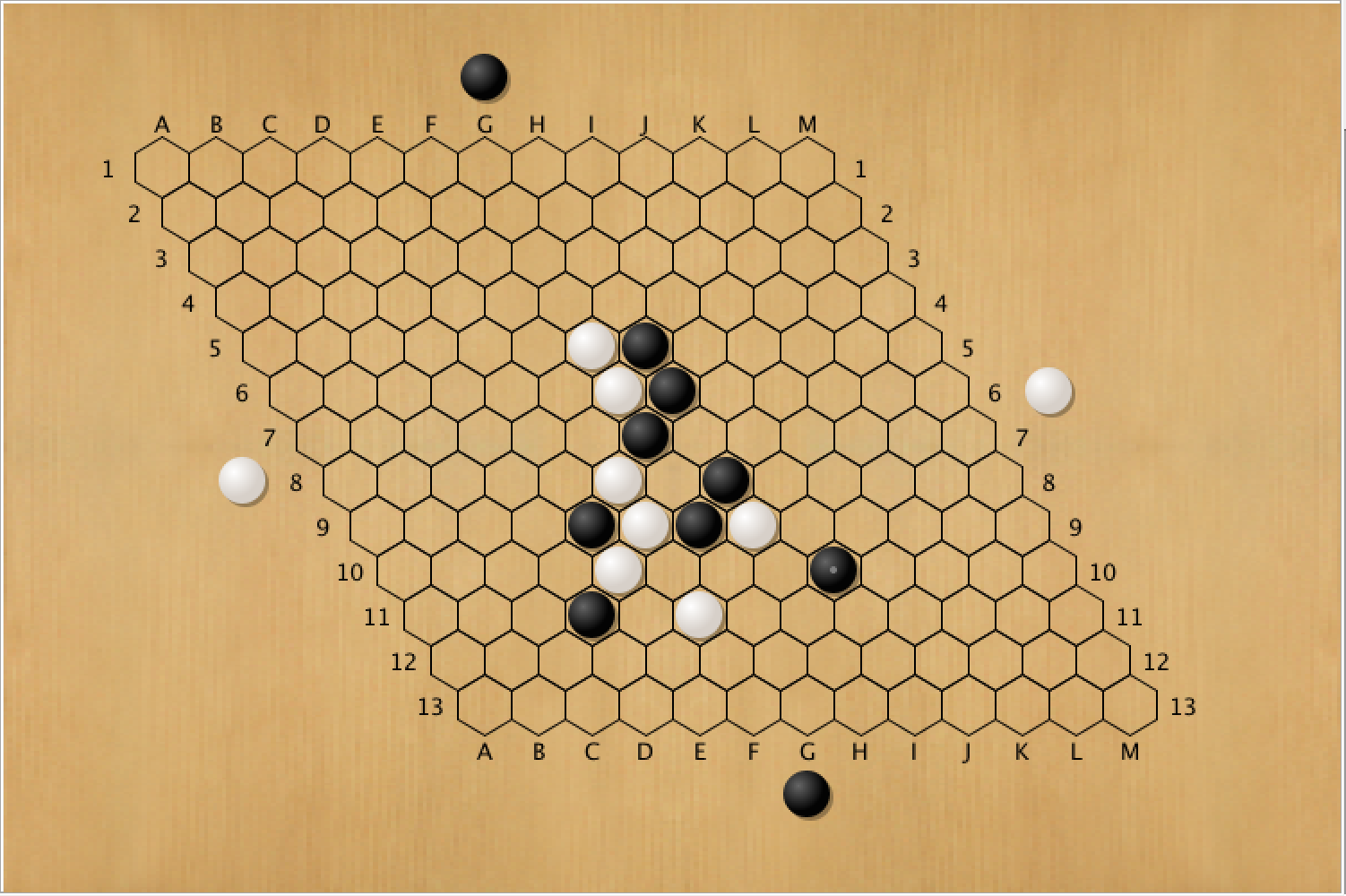}
  \caption{A hex game in progress. Black wants to join top and bottom, White wants to join left and right.}
  \label{fig:hex1}
\end{subfigure}\hfill
\begin{subfigure}[t]{.45\textwidth}
  \centering
      \includegraphics[width=1\textwidth]{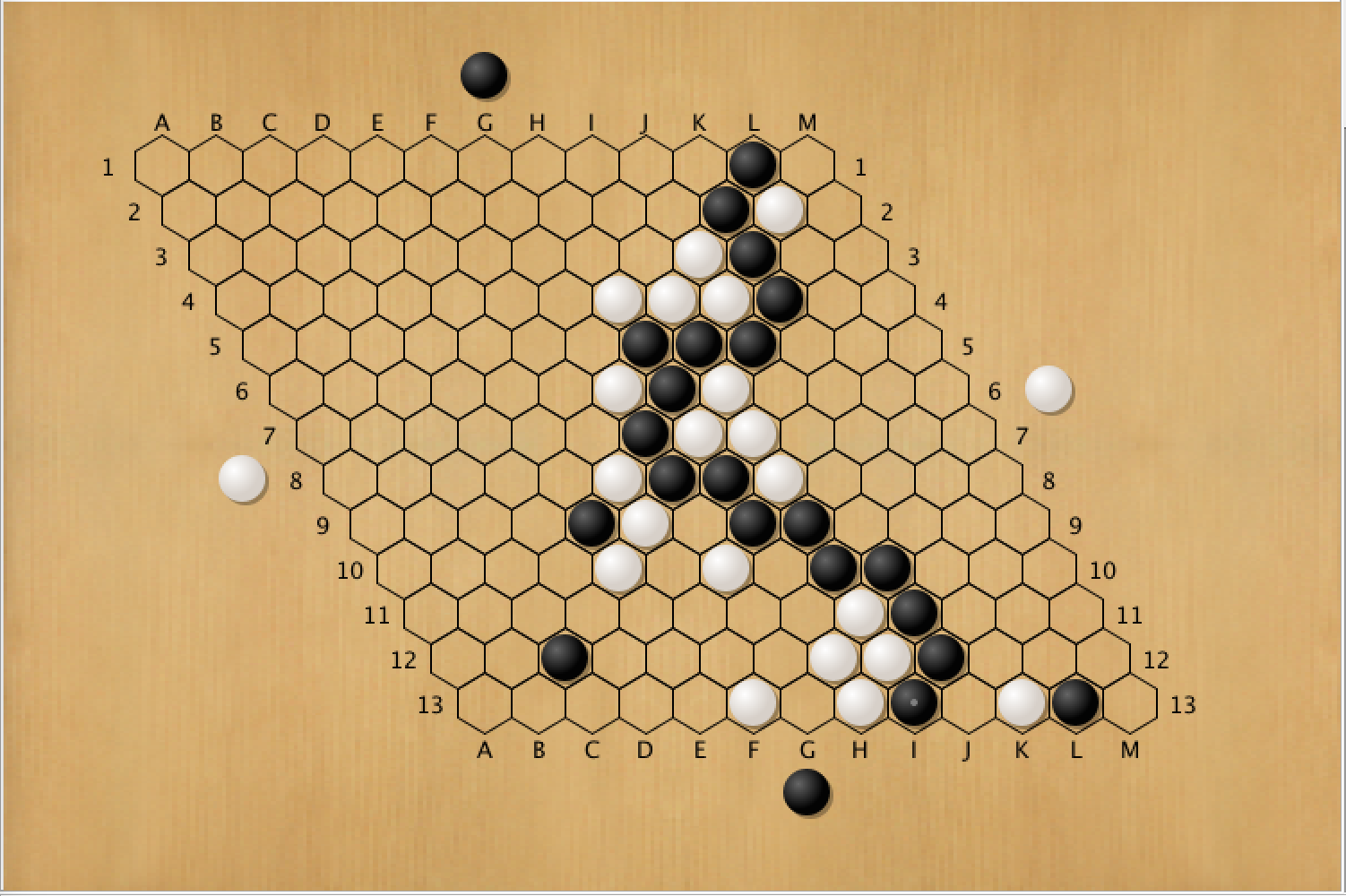}
  \caption{A finished hex game. Black wins.}
  \label{fig:hex2}
\end{subfigure}
\caption{The game of hex.}
\label{fig:hex}
\end{figure}

\subsection{Related Work}
The two works that inspire this paper
are \cite{mnih-dqn-2015} and \cite{Silver2016}, both from Google DeepMind. 

\cite{mnih-dqn-2015} introduces Deep Q-learning with Experience Replay. 
Q-learning is a reinforcement learning (RL) algorithm that learns a mapping from states to action values by backing up action value estimates from subsequent states to improve those in previous states. 
In Deep Q-learning the mapping from states to action values is learned by a Deep Neural network. 
Experience replay extends standard Q-learning by storing agent experiences in a memory buffer and sampling from these experiences every time-step to perform updates. 
This algorithm achieved superhuman performance on a several classic Atari games using only raw visual input. 

\cite{Silver2016} introduces AlphaGo, a Go playing program that combines Monte Carlo tree search with convolutional neural networks: 
one guides the search (policy network), another evaluates position quality (value network). 
Deep reinforcement learning (RL) is used to train both the value and policy networks, which each take a representation of the gamestate as input. The policy network outputs a probability distribution over available moves indicating the likelihood of choosing each move. 
The value network outputs a single scalar value estimating $V(S)=P(win|S)-P(loss|S)$, the expected win probability minus the expected loss probability for the current boardstate $S$. 
Before applying RL, AlphaGo's network training begins with supervised mentoring: the policy network is trained to replicate moves from a database of human games.
Then the policy network is trained by playing full games against past versions of their network, followed by increasing the probability of moves played by the winner and decreasing the probability of moves played by the loser. 
Finally the value network is trained by playing full games from various positions using the trained policy network, and performing a gradient descent update based on the observed game outcome. 
Temporal difference (TD) methods --- which update value estimates for previous states based on the systems own evaluation of subsequent states, rather than waiting for the true outcome --- are not used.

An early example of applying RL with a neural network to games is TD-gammon \cite{tesauro1995temporal}. There a network trained with TD methods to approximate state values achieved superhuman play. Recent advances in deep learning have opened the doors to apply such methods to more games.

\subsection{Overview of This Work}
In this work we explore the application of Deep Q-learning with Experience Replay, introduced in \cite{mnih-dqn-2015}, to Hex. There are several challenges involved in applying this method, so successful with Atari,
to Hex. One challenge is that there are fewer available actions in Atari than in Hex (e.g.\ there are 169 possible initial moves in 13$\times$13 Hex). Since Q-learning performs a maximization over all available actions, this large number might cause cause the noise in estimation to overwhelm the useful signal, resulting in catastrophic maximization bias. However in our work we found the use of a convolutional neural network ---
which by design learns features that generalize over spatial location ---
yielded good results.

Another challenge is that the reward signal in Hex occurs only at the end of a game, so (with respect to move actions) is infrequent, meaning that most updates are based only on network evaluations without immediate win/loss feedback. The question is whether the learning process will allow this end-of-game reward information to propagate back to the middle and early game. To address this challenge, we use supervised mentoring, training the network first to replicate the action values produced by a heuristic over a database of positions. Such training is faster than RL, and allows the middle and early game updates to be meaningful at the start of Q-learning, without having to rely on end-of-game reward propagating back from the endgame. As with AlphaGo \cite{Silver2016}, we apply this heuristic only to initialize the network: the reward in our Q-learning is based only on the outcome of the game being played.

The main advantage of using a TD method such as Q-learning over training based only on final game outcomes, as was done with AlphaGo, is data efficiency. Making use of subsequent evaluations by our network allows the system to determine which specific actions are better or worse than expected based on previous training by observing where there is a sudden rise or fall in evaluation. A system that uses only the final outcome can only know that the set of moves made by the winner should be encouraged and those made by the loser discouraged, even though many of the individual moves made by the winner may be bad and many of the individual moves made by the loser may be good. We believe this difference is part of what allows us to obtain promising results using less computing power than AlphaGo.

\subsection{Reinforcement Learning}
Reinforcement learning is a process that learns from actions that lead to a goal. An agent learns from the environment and makes decisions. Everything that the agent can interact with is called the environment. The agent and environment interact continually: the agent selecting actions and the environment responding to those actions and presenting new situations to the agent. The environment also reports rewards: numerical/scalar values that the agent tries to maximize over time. A complete specification of an environment defines a task, which is one instance of the reinforcement learning problem. 

\begin{figure}[!ht]
\centering
 \includegraphics[width=0.65\textwidth]{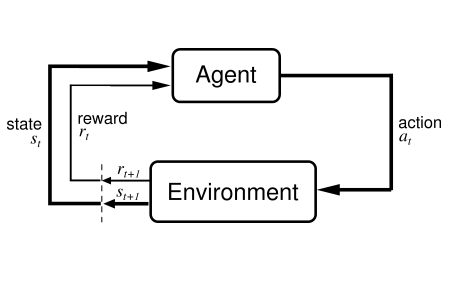}
  \caption{The agent-environment interaction in reinforcement learning.}
  \label{fig:RL}
\end{figure}

The agent and environment interact at each discrete time step (t = 0,1,2,3...). At each time step the agent receives some representation of the environment's state, $s_{t}$  $\epsilon$  $\mathcal{S}$, where $\mathcal{S}$ is the set of possible states, and on that basis selects an action, $a_{t}$ $\epsilon$ $\mathcal{A}_{t} $ , where $\mathcal{A}_{t}$ is the set of actions available in state $s_{t}$. One time step later, in part as a consequence of its action, the agent receives a numerical reward, $r_{t+1}$ $\epsilon$ $\mathcal{R}$ , and finds itself in a new state $s_{t+1}$.

The purpose or goal of the agent is formalized in terms of a special reward signal passing from the environment to the agent. At each time step, the reward is a scalar, $r_{t}$ $\epsilon$ $\mathcal{R} $ . Informally, the agent's goal is to maximize the total amount of reward it receives. This means maximizing not immediate reward, but cumulative reward in the long run. The field of reinforcement learning is primarily the study of methods for tackling  this challenge.

A RL agent chooses actions according to a policy $\pi(a|s)$ which is a probability distribution over all possible actions for the current state. The policy may be deterministic or stochastic. For a given policy we define the value of a state $v_\pi(S)$ as the expectation value of cumulative reward from state S if we follow $\pi$. 

\[ v_\pi(S) = E_\pi(\sum_{t=1}^\infty \gamma^{t-1} r_t | s_0=S) \]

where $\gamma$ is a discount factor indicating how much more to credit immediate reward than long term reward; this is generally necessary to ensure reward is finite if the agent-environment interaction continues indefinitely, however it may be omitted if the interaction ends in bounded time, for example in a game of Hex. For a given problem we define the optimal policy $\pi_*$ (not necessarily unique) as that which produces the highest value in every state. We then denote this highest achievable value as $v_*(S)$. Note that neither $v_\pi(S)$ or $v_*(S)$ are tractable to compute in general, however it is the task of a wide variety of RL algorithms to estimate them from the agent's experience.

Similarly we can define for any policy the action value of each state action pair $q_\pi(S,A)$ which, analogous to $v_\pi(S)$, is defined as the expectation value of cumulative reward from state S if we take action A and follow $\pi$ after that. Similarly we can define $q_*(S,A)$ as $q_{\pi_*}(S,A)$. Notice that choosing the action with the highest $q_*(S,A)$ in each state is equivalent to following the optimal policy $\pi_*$.

See \cite{Sutton1998} for excellent coverage of these topics and many others pertaining to RL.

\subsection{Deep Q-learning}
Q-learning is based on the following recursive expression, called a Bellman equation for $q_*(S,A)$.

\[ q_*(s_t,a_t) = E(r_{t+1} + \gamma\underset{a}{max}\ q_*(s_{t+1},a) | s_t,a_t)\]

Note that this expression can be derived from the definition of $q_*(s_t,a_t)$. From this expression we formulate an update rule which allows us to iteratively update an estimate of $q_*(S,A)$, typically written $Q_*(S,A)$ or simply $Q(S,A)$ from the agents stream of experience as follows:

\[Q(s_t,a_t) \overset{\alpha}{\leftarrow} r_{t+1} + \gamma\underset{a}{max}\ Q(s_{t+1},a)\]

Where in the tabular case (all state action pairs estimated independently)  ``$\overset{\alpha}\leftarrow$'' would represent moving the left-hand-side value toward the right-hand-side value by some step size $\alpha$ fraction of the total difference, in the function approximation case (for example using a neural network) we use it to represent a gradient descent step on the left value decreasing (for example) the squared difference between them. Since a maximization is required, if the network for Q were formulated as a map directly from state-action pairs to values, it would be necessary to perform one pass through the network for each action in each timestep. Because this would be terribly inefficient (particularly in the case of Hex which has up to 169 possible actions) and also because action values for a given state are highly correlated, we instead follow \cite{mnih-dqn-2015} and use a network that outputs values for all actions in one pass.

Note that since we take the maximum over the actions in each state, it is not necessary to actually follow the optimal policy to learn the optimal action values, though we do need to have some probability to take each action in the optimal policy. If the overlap with the policy followed and the optimal policy is greater we will generally learn faster. Usually the policy used is called epsilon-greedy which takes the action with the highest current $Q(s_t,a_t)$ estimate most of the time but chooses an action at random some fraction of the time. This method of exploration is far from ideal and improving on it is an interesting area of research in modern RL.

Having learned an approximation of the optimal action values, at test time we can simply pick the highest estimated action value in each state, and hopefully in doing so follow a policy that is in some sense close to optimal.

%------------------------------------------- Implementation -------------------------------------%
\section{Method}
\subsection{Problem Structure}
We use the convention that a win is worth a reward of +1 and a loss is worth -1.  All moves that do not end the game are worth 0. We are in the episodic case, meaning that we wish to maximize our total reward over an episode (i.e. we want to win and not lose), hence we use no discounting ($\gamma=1$). Note that the ground truth $q_*(S,A)$ value is either 1 or -1 for every possible state-action pair (the game is deterministic and no draws are possible, hence assuming perfect play one player is certain to lose and the other is certain to win). The network's estimated $q_*(S,A)$ value Q(S)[A] then has the interpretation of subjective probability that a particular move is a win, minus the subjective probability that it is a loss (roughly speaking $Q(S)[A]=P(win|S,A)-P(loss|S,A)$). We seek to predict the true value of $q_*(S,A)$ as accurately as possible over the states encountered in ordinary play, and in doing so we hope to achieve strong play at test time by following the policy which takes action $\underset{a}{argmax}\ Q(s)[a]$, i.e. the move with highest estimated win probability, in each state.

\subsection{State Representation}
The state of the Hex board is encoded as a 3 dimensional array with 2 spatial dimensions and 6 channels as follows:
%\begin{itemize}
%\item white stone present
%\item black stone present
%\item white stone group connected to left edge
%\item white stone group connected to right edge
%\item black stone group connected to top edge
%\item black stone group connected to bottom edge
%\end{itemize}
white stone present; ~
black stone present; ~
white stone group connected to left edge; ~
white stone group connected to right edge; ~
black stone group connected to top edge; ~
black stone group connected to bottom edge.

\begin{figure}[!ht]
\centering
      \includegraphics[width=0.65\textwidth]{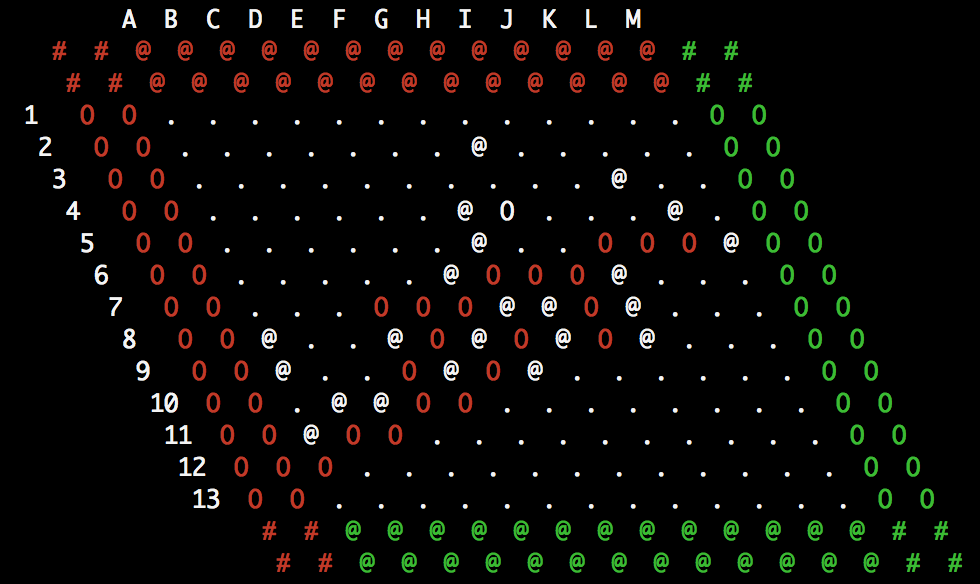}
  \caption{A visualization of the board representation fed to NeuroHex. O is white, @ is black, red is north or east edge connected depending on the color of the associated stone, similarly green is south or west edge connected. Note that though the board is actually 13 by 13 the input size is 17 by 17 to include 2 layers of padding to represent the board edge and allow placement of 5 by 5 filters along it. Cells in the corners marked with \# are uniquely colored both white and black from the perspective of the network.}
\label{fig:input}
\end{figure}

In addition to the 13 by 13 Hex board, the input includes 2 cells of padding on each side which are connected to the corresponding edge by default and belong to the player who is trying to connect to that edge. This padding serves a dual purpose of telling the network where the edges are located and allowing both 3 by 3 and 5 by 5 filters to be placed directly on the board edge without going out of bounds.
We note that AlphaGo used a far more feature rich input representation including a good deal of tactical information specific to the game of Go. Augmenting our input representation with additional information of this kind for Hex could be an interesting area of future investigation. Our input format is visualized in figure \ref{fig:input}.

\subsection{Model}
We use the Theano library \cite{bergstra+al:2010-scipy}\cite{Bastien-Theano-2012} to build and train our network. Our network architecture is inspired by that used by Google DeepMind for AlphaGo's policy network \cite{Silver2016}. Our network consists of 10 convolutional layers followed by one fully connected output layer. The AlphaGo architecture was fully convolutional, with no fully connected layers at all in their policy network, although the final layer uses a 1x1 convolution. We however decided to employ one fully connected layer, as we suspect that this architecture will work better for a game in which a global property (i.e., in Hex, have you connected your two sides?) matters more than the sum of many local properties (i.e., in Go, which local battles have you won?). For future work, it would of interest to explore the effect of the final fully connected layer in our architecture.
\begin{figure}[!ht]
\centering
      \includegraphics[width=0.65\textwidth]{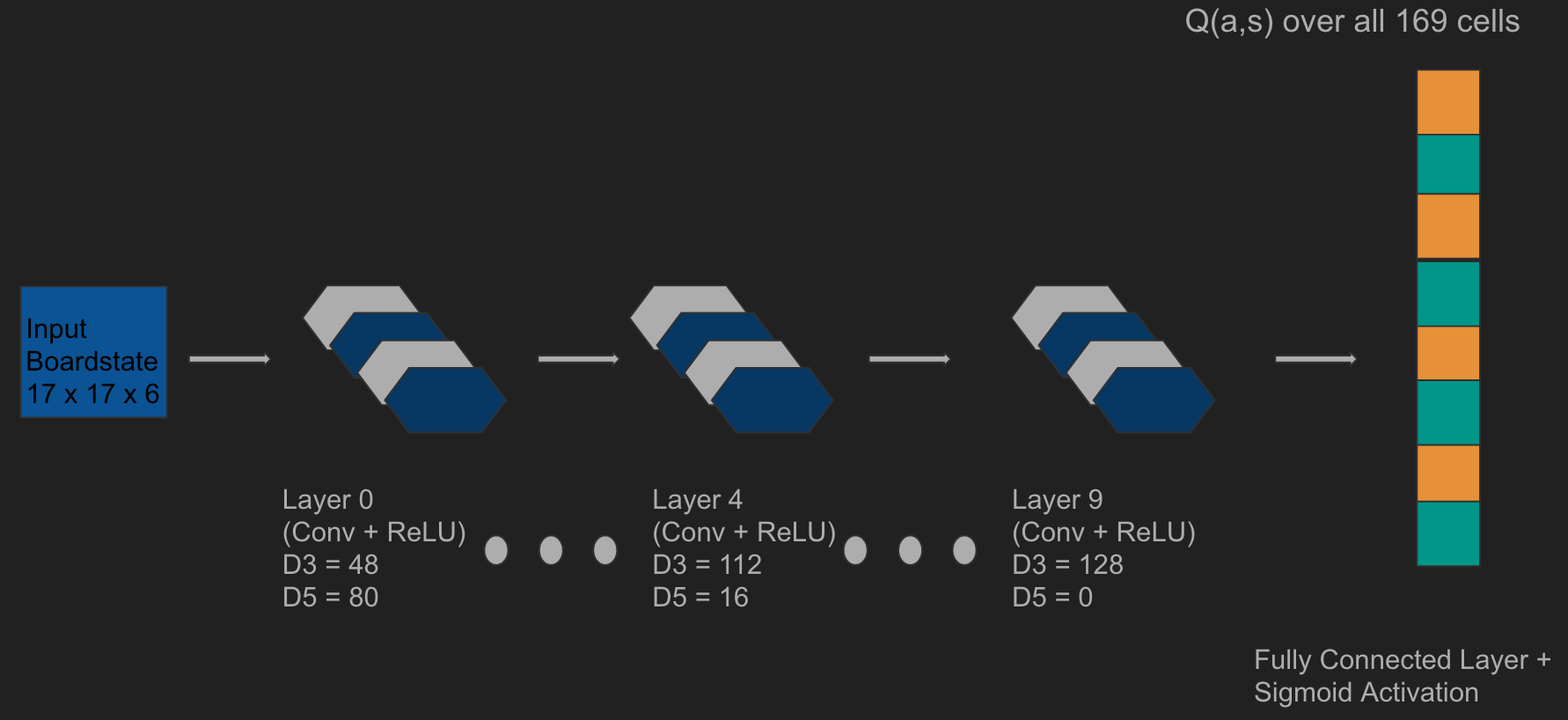}
  \caption{Diagram showing our our network layout, D3 indicates number of diameter 3 filters, D5 indicates number of diameter 5 filters in the layers shown.}
\label{fig:model}
\end{figure}

Filters used are hexagonal rather than square to better capture the different notion of locality in the game of Hex. Hexagonal filters were produced simply by zeroing out appropriate elements of standard square filters and applying Theano's standard convolution operation. Each convolutional layer has a total of 128 filters which consist of a mixture of diameter 3 and diameter 5 hexagons, all outputs use stride 1 and are 0 padded up to the size of the original input. The output of each convolutional layer is simply the concatenation of the padded diameter 5 and diameter 3 outputs. All activation function are Rectified Linear Units (ReLU) with the exception of the output which uses $1-2\sigma(x)$ (a sigmoid function) in order to obtain the correct range of possible action values. The output of the network is a vector of action values corresponding to each of the board locations. Unplayable moves (occupied cells) are still evaluated by the network but simply ignored where relevant since they are trivially pruned.

\subsection{Position Database}
While it would be possible to train the network purely by Q-learning from self play starting from the empty board every game, we instead generated a database of starting positions from 10,000 games played by a noisy version of a strong hex playing program based on alpha-beta search called Wolve \cite{arneson2008wolve}. To generate this database each game was started with a random move and each subsequent move was chosen by a softmax over Wolve's move evaluations to add additional variability. These positions were used for two separate purposes. First they were used in mentoring (supervised learning to provide a reasonable initialization of the network before Q-learning) which is described in the section below. Second to randomly draw a starting position for each episode of self play by the network. This second usage was meant to ensure that the network experiences a wide variety of plausible game positions during training, without going through the potentially arduous process of finding this variety by random exploration. In the future it could be interesting to see how important these two things are to the success of the algorithm. 

\subsection{Mentoring}
Before beginning Q-learning, the network was trained to replicate (by stochasic gradient descent on the mean squared error) the action values produced by a variant of a common hex heuristic based on electrical resistance\cite{anshelevich2000game}, over the position database discussed in the previous section. The idea of the heuristic is to place a voltage drop across the two edges a player is trying to connect, then take the players own cells to be perfect conductors, opponent cells to be perfect insulators, and empty cells to be finite resistors. The usual version of the heuristic then computes a score (an arbitrary positive real number with no statistical interpretation) of the position as the ratio of current traveling across the board for each player. Because we wanted instead to generate heuristic action values between -1 and 1 for each move, it was necessary to modify this heuristic. We did this by computing estimates of the current across the board $C_1'(a)$ and $C_2'(a)$ for the player to move and their opponent respectively following the player to move playing into cell $a$ (the true value could have been used by playing each move and recomputing the current, but we use a simple estimate based on the current through each cell to save time). The action value of a cell was then taken to be:
\begin{equation*}
Q(a) = \begin{cases} 1-C_2'(a)/C_1'(a), & \mbox{if } C_1'(a)>C_2'(a) \\ C_1'(a)/C_2'(a) - 1, & \mbox{if } C_2'(a)>C_1'(a) \end{cases}
\end{equation*}
In any case the details here are not terrible important and similar results could have likely been obtained with a simpler heuristic. The important thing is that the heuristic supervised pre-training gives the network some rough initial notion that the goal of the game is to connect the two sides. This serves the primary purpose of skipping the potentially very long period of training time where most updates are meaningless since the reward signal is only available at the end of an episode and every other update is mostly just backing up randomly initialized weights. It also presumably gives the network an initial set of filters which have some value in reasoning about the game. Note that the final trained network is much stronger than this initialization heuristic.

\subsection{Q-learning Algorithm}
We use Deep Q-learning with experience replay in a manner similar to Google DeepMind's Atari learning program \cite{mnih-dqn-2015}. Experience replay means that instead of simply performing one update at a time based on the last experience, we save a large set of the most recent experiences (in our case 100,000), and perform a random batch update (batch size 64) drawn from that set. This has a number of benefits including better data efficiency, since each experience is sampled many times and each update uses many experiences; and less correlation among sequential updates. We use RMSProp \cite{Tieleman2012} as our gradient descent method.

One notable difference between our method and \cite{mnih-dqn-2015} is in the computation of the target value for each Q-update. Since in the Atari environment they have an agent interacting with an environment (an Atari game) rather than an adversary they use the update rule $Q(s_t,a_t)\overset{\alpha}{\leftarrow} r_{t+1}+\gamma \underset{a}{max}\ Q(s_{t+1},a)$, where again we use $\overset{\alpha}{\leftarrow}$ to indicate the network output on the left is moved toward the target on the right by a gradient descent update to reduce the squared difference. Here $\gamma$ is a discount factor between 0 and 1 indicating how much we care about immediate reward v.s. long-term reward. 

In our case the agent interacts with an adversary who chooses the action taken in every second state, we use the following gradient descent update rule: $Q(s_t,a_t)\overset{\alpha}{\leftarrow} r_{t+1}- \underset{a}{max}\ Q(s_{t+1},a)$. Note that we take the value to always be given from the perspective of the player to move. Thus the given update rule corresponds to stepping the value of the chosen move toward the negation of the value of the opponents next state (plus a reward, nonzero in this case only if the action ends the game). This update rule works because with the way our reward is defined the game is zero-sum, thus the value of a state to our opponent must be precisely the negation of the value of that state to us. Also in our case we are not concerned with how many moves it takes to win and we suspect using a discount factor would only serve to muddy the reward signal so we set $\gamma=1$.

Our network is trained only to play as white, to play as black we simply transform the state into an equivalent one with white to play by transposing the board and swapping the role of the colors. We did it this way so that the network would have less to learn and could make better use of its capacity. An alternative scheme like outputting moves for both white and black in each state seems wasteful as playing as either color is functionally the same (ignoring conventional choices of who plays first). It is however an interesting question whether training the network to pick moves for each color could provide some useful regularization.

Our Q-learning algorithm is shown in algorithm \ref{alg:DQL}. Note that we include some forms of data augmentation in the form of randomly flipping initial states to ones that are game theoretically equivalent by symmetry, as well as randomly choosing who is to move first for the initial state (irrespective of the true player to move for a given position). The latter augmentation will result in some significantly imbalanced positions since each move can be crucial in hex and losing a move will generally be devastating. However since our player is starting off with very little knowledge of the game, having these imbalanced positions where one player has the opportunity to exploit serious weakness presumably allows the network to make early progress in identifying simple strategies. A form of curriculum learning where these easier positions are trained on first followed by more balanced positions later could be useful, but we did not investigate this here. We also flip the state resulting from each move to an equivalent state with 50\% probability, a form of game specific regularization to capture a symmetry of the game and help smooth out any orientation specific noise generated in training.

\begin{algorithm}[!ht]
initialize replay memory $M$, Q-network $Q$, and state set $D$\\
\For{desired number of games}{
$s$ = position drawn from $D$\\
randomly choose who moves first\\
randomly flip $s$ with 50\% probability\\
 \While{game is not over}{
  $a$ = epsilon\_greedy\_policy($s$, $Q$)\\
  $s_{next}$ = $s$.play($a$)\\
  \eIf{game is over}{
  	$r$=1\\
  }{
  	$r$=0\\
  }
  randomly flip $s_{next}$ with 50\% probability\\
  $M$.add\_entry(($s$,$a$,$r$,$s_{next}$))\\
  ($s_t$,$a_t$,$r_{t+1}$,$s_{t+1}$) = $M$.sample\_batch()\\
  $target_t$ = $r_{t+1}-\underset{a}{max}\ Q(s_{t+1})[a]$\\
  Perform gradient descent step on Q to reduce $(Q(s_t)[a_t]-target_t)^2$\\
  s = $s$.play($a$)
  }
}
\caption{Our Deep Q-learning algorithm for hex. epsilon\_greedy\_policy($s$, $Q$) picks the action with the highest Q value in s 90\% of the time and 10\% of the time takes a random action to facilitate exploration. $M$.sample\_batch() randomly draws a mini-batch from the replay memory. Note that in two places we flip states (rotate the board 180$^{\circ}$) at random to capture the symmetry of the game and mitigate any orientation bias in the starting positions.}
\label{alg:DQL}
\end{algorithm}

%------------------------------------------- Results------- -------------------------------------%
\section{Results}
\begin{figure}[!ht]
	\centering
	\includegraphics[width=0.9\textwidth]{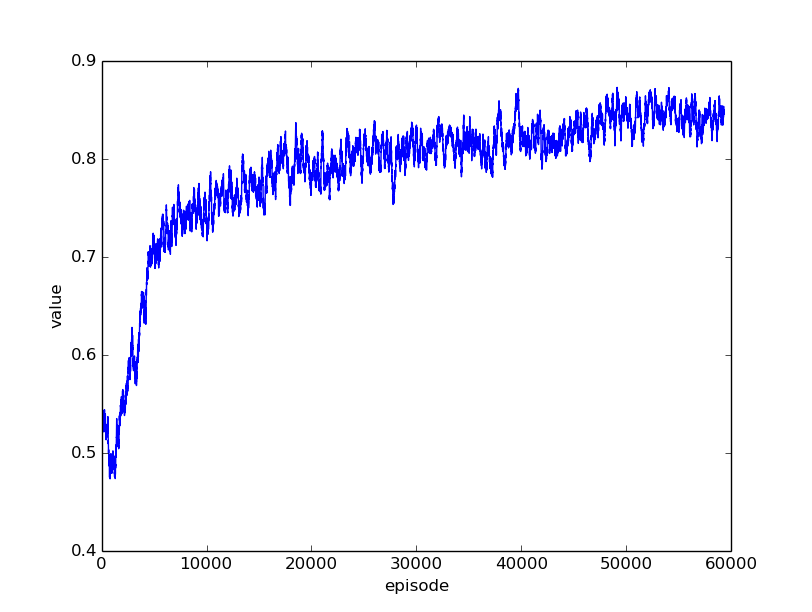}
	\caption{Running mean (over 200 epsiodes) of the magnitude (absolute value) of  the value (max over all action values from that position) for positions encountered in training. For each position, its ground truth value is either -1 or 1, so this graph indicates network's confidence in its evaluation of positions that it encounters in training.}
	\label{fig:value_plot}
\end{figure}

\begin{figure}[!ht]
	\centering
	\includegraphics[width=0.9\textwidth]{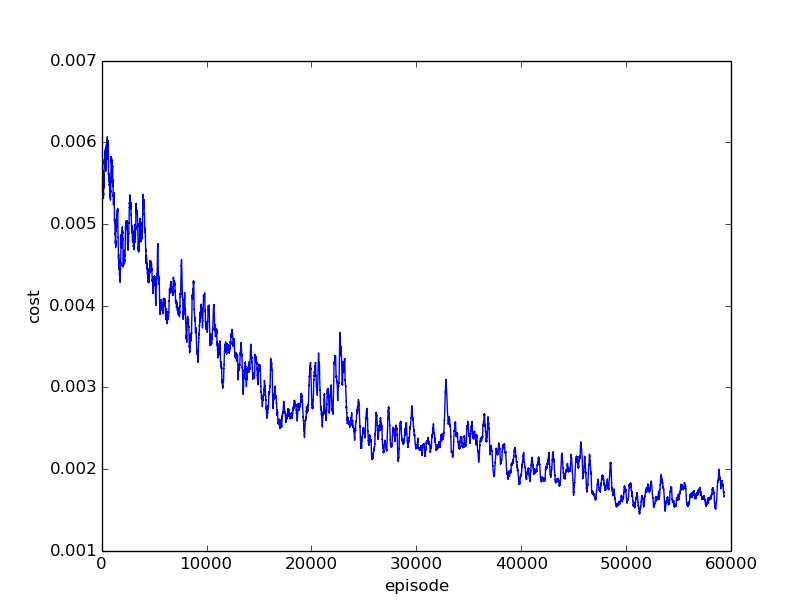}
	\caption{Running mean (over 200 episodes) of average cost of updates in Algorithm~\ref{alg:DQL}: squared difference between current Q-value and target (max Q-value of next position). So this graph indicates the rate at which the network is learning.}
  \label{fig:cost_plot}
\end{figure}

To measure the effectiveness of our approach, we measure NeuroHex's
playing strength, rate of learning, and stability of learning.
Our results are summarized in
Figures \ref{fig:winrate}, \ref{fig:value_plot}, and \ref{fig:cost_plot} respectively. 

\begin{figure}[!ht]
\centering
\begin{tabular}{|c|c|c|c|}  \hline 
first move & MoHex time/move & NeuroHex black & NeuroHex white \\ \hline
unrestricted & 1 & .20 & .02 \\ \hline
all 169 openings & 1 & .11& .05 \\ \hline
all 169 openings & 3 & .09 & .02 \\ \hline
all 169 openings & 9 & .07 & .01 \\ \hline
\end{tabular}
\caption{NeuroHex v MoHex win rates. Black is 1st-player. The unrestricted winrates are over 1000 games, the others are over 169 games.}
\label{fig:winrate}
\end{figure}

Figure \ref{fig:value_plot} shows the average magnitude of maximal action value output by the network.
Figure \ref{fig:cost_plot} shows the average cost for each Q-update performed during training as a function of episode number. Both of these are internal measures, indicating the reasonable convergence and stability of the training procedure; they do not however say anything about the success of the process in terms of actually producing a strong player. After an initial period of rapid improvement, visible in each plot, learning appears to become slower. Interestingly the cost plot seems to show a step-wise pattern, with repeated plateaus followed by sudden drops. This is likely indicative of the effect of random exploration, the network converges to a set of evaluations that are locally stable, until it manages to stumble upon, through random moves or simple the right sequence of random batch updates, some key feature that allows it to make further improvement. Training of the final network discussed here proceeded for around 60,000 episodes on a GTX 970 requiring a wall clock time of roughly 2 weeks. Note that there is no clear indication in the included plots that further improvement is not possible, it is simply a matter of limited time and diminishing returns.

\begin{figure}[!ht]
\centering
\begin{subfigure}[t]{.5\textwidth}
  \centering
  \includegraphics[width=1\textwidth]{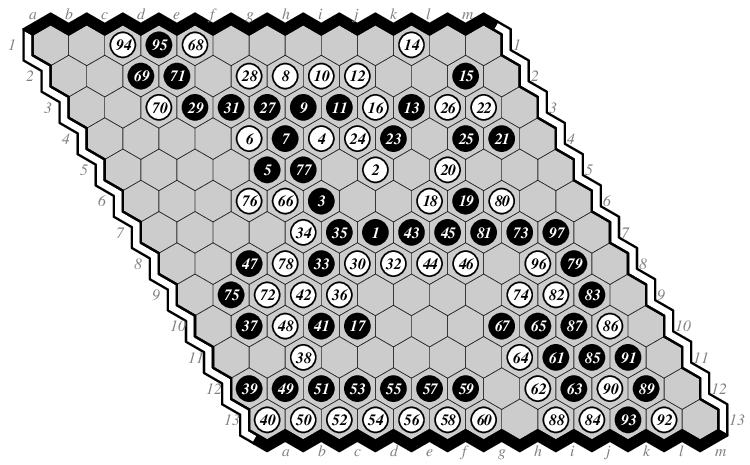}
  \caption{Win as black.}
  \label{fig:black_win}
\end{subfigure}\hfill
\begin{subfigure}[t]{.5\textwidth}
  \centering
      \includegraphics[width=1\textwidth]{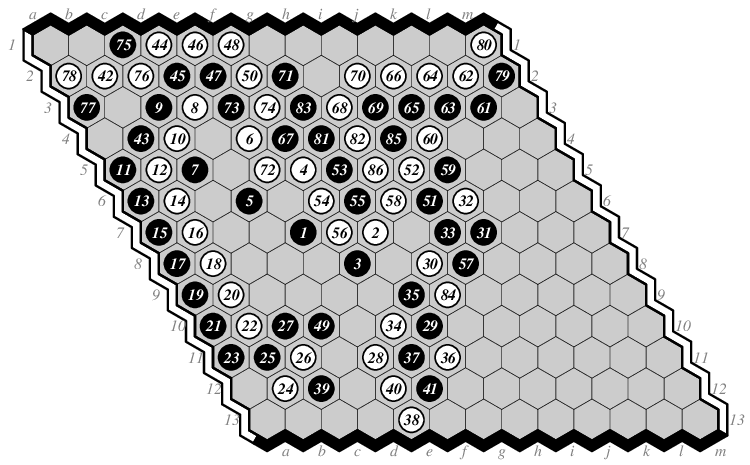}
  \caption{Win as white.}
  \label{fig:white_win}
\end{subfigure}
\caption{Example wins for NeuroHex over MoHex.}
\label{fig:wins}
\end{figure}

\begin{figure}[!ht]
\centering
\begin{subfigure}[t]{.5\textwidth}
  \centering
  \includegraphics[width=1\textwidth]{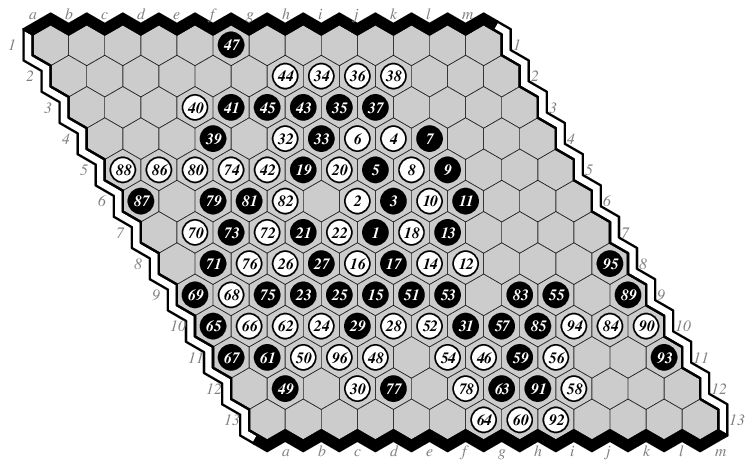}
  \caption{Loss as black.}
  \label{fig:black_loss}
\end{subfigure}\hfill
\begin{subfigure}[t]{.5\textwidth}
  \centering
      \includegraphics[width=1\textwidth]{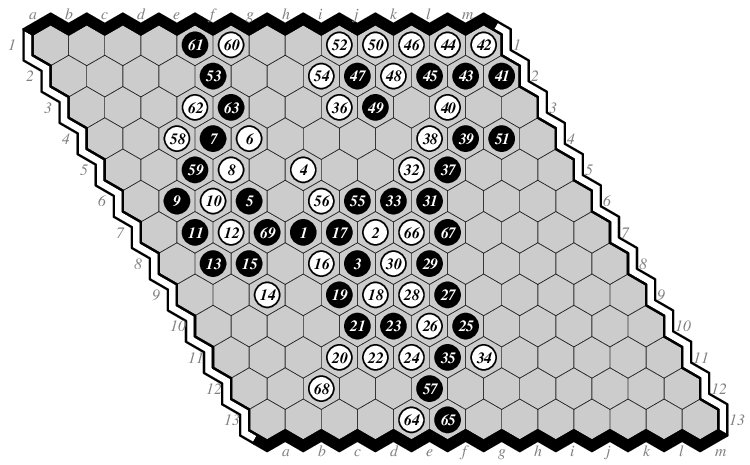}
  \caption{Loss as white.}
  \label{fig:white_loss}
\end{subfigure}
\caption{Example wins for MoHex over NeuroHex.}
\label{fig:losses}
\end{figure}

We evaluate NeuroHex by testing against the Monte-Carlo tree search player MoHex\cite{MoHex2}\cite{AHH10b}, currently the world's strongest hexbot. See Figure~\ref{fig:winrate}.
Four sample games are shown in Figures  \ref{fig:wins} and \ref{fig:losses}. On board sizes up to at least 13$\times$13 there is a significant first-player advantage. To mitigate this, games are commonly played with a ``swap rule'' whereby after the first player moves the second player can elect either to swap places with the first player by taking that move and color or to continue play normally. Here, we mitigate this first-player advantage by running two kinds of tournaments: in one, in each game the first move is unrestricted; in the other, we have rounds of 169 games, where the 169 first moves cover all the moves of the board. As expected, the all-openings win-rates lie between the unrestricted 1st-player and 2nd-player winrates. To test how winrate varies with MoHex move search time, we ran the all-openings experiment with 1s, 3s, 9s times.

MoHex is a highly optimized C++ program. Also, in addition to Monte Carlo tree search, it uses makes many theorems for move pruning and early win detection. So the fact that NeuroHex, with no search, achieves a non-zero success rate against MoHex,
even when MoHex plays first, is remarkable.


%-------------------------------------- Conclusion ------------------------------------------%
\section{Discussion and Conclusion}
The DeepMind authors showed that
Deep RL based on final outcomes can be used to build a superhuman Gobot; we have shown that Q-learning can be used to build a strong Hexbot. Go and Hex have many similarities --- two-player alternate turn game on a planar board in which connectivity is a key feature --- and we expect that our methods will apply to Go and many other similar games. 

Before our work we wondered whether the large action spaces that arise in Hex would result in the temporal difference updates of Q-learning being overwhelmed by noise. We also wondered whether the assignment of correct credit for actions would be feasible
in light of the paucity of rewards among the large number of states. But the win-rate of NeuroHex against
the expert-level player MoHex after training only two weeks suggests that the generalization ability of convolutional neural networks together with the use of supervised mentoring is sufficient to overcome these challenges. One property of Hex that might have contributed to the success of our methods is that --- unlike Go --- it is an all-or-nothing game. In particular, Hex has a ``sudden death'' property: in many positions, for most moves, it is easy to learn that those moves lose. In such positions it is comparatively easy task for the network to learn to distinguish the few good moves from the many bad ones.

In light of supervised mentoring one could ask to what extent our training is using the reward signal at all, versus simply back-propagating the heuristic initialization values. We would like to address this question in the future, for example by testing the procedure without supervised mentoring, although this might not be important from the perspective of building a working system. If the heuristic is good then many of the values should already be close to the true value they would eventually converge to in Q-learning. Assuming, as is often the case, that heuristic values near the endgame are better than those near the start, we will be able to perform meaningful backups without directly using the reward signal. To the extent the heuristic is incorrect it will eventually be washed out by the reward signal --- the true outcome of the game --- although this may take a long time.

We suspect that our network would show further improvement with further training, although we have no idea to what extent. We also suspect that incorporating our network into a player such as MoHex, for example to bias the initial tree search, would strengthen the player.

%------------------------------------------- Future Work -------------------------------------%
\section{Future Work}
Throughout the paper we have touched on possible directions for further research. Here are some possibilities: augment the input space with tactical or strategic features (e.g. in Hex, virtual connections, dead cells and capture patterns);
build a search based player using the trained network, or incorporate it into an existing search player (e.g. MoHex); determine the effectiveness of using an actor-critic method to train a policy network along side the Q network to test the limits of the learning process;
 find the limits of our approach by training for a longer period of time;
 determine whether it is necessary to draw initial positions from a state database, or to initialize with supervised learning; investigate how using a fully convolutional neural network compares to the network with one fully connected layer we used.

%------------------------------------------- References ----------------------------------------%

\bibliography{bibliography}
\bibliographystyle{plain}

\end{document}